\documentclass{article}

 \usepackage{graphicx} 
 \usepackage{enumitem} 

\usepackage[preprint, nonatbib]{neurips_2024}

\usepackage{amssymb}
\usepackage{diagbox}
\usepackage[utf8]{inputenc} 
\usepackage[T1]{fontenc}    
\usepackage[colorlinks,
linkcolor=purple,
anchorcolor=cyan,
citecolor=blue,
]{hyperref}       
\usepackage{amsmath}
\usepackage{url}            
\usepackage{booktabs}       
\usepackage{amsfonts}       
\usepackage{nicefrac}       
\usepackage{microtype}      
\usepackage{multirow}
\usepackage{ulem}
\usepackage{anyfontsize}
\usepackage{hyperref}
\usepackage{textcomp}
 \usepackage{arydshln}
 \usepackage[dvipsnames]{xcolor}

\newcommand{\nocomment}{}

\ifx\nocomment\undefined
 \NewDocumentCommand{\heng}
    { mO{} }{\textcolor{red}{\textsuperscript{\textit{Heng}}\textsf{\textbf{\small[#1]}}}}
 \NewDocumentCommand{\pengfei}
    { mO{} }{\textcolor{purple}{\textsuperscript{\textit{Pengfei}}\textsf{\textbf{\small[#1]}}}}

 \NewDocumentCommand{\zoey}
    { mO{} }{\textcolor{blue}{\textsuperscript{\textit{Zoey}}\textsf{\textbf{\small[#1]}}}}

 \NewDocumentCommand{\yuji}
    { mO{} }{\textcolor{cyan}{\textsuperscript{\textit{yuji}}\textsf{\textbf{\small[#1]}}}}

 \NewDocumentCommand{\yi}
    { mO{} }{\textcolor{green}{\textsuperscript{\textit{Yi}\textsf{\textbf{\small[#1]}}}}}

\NewDocumentCommand{\jiateng}
    { mO{} }{\textcolor{orange}{\textsuperscript{\textit{Jiateng}\textsf{\textbf{\small[#1]}}}}}

\else
    \NewDocumentCommand{\heng}
    { mO{} }{\textcolor{red}{}}
     \NewDocumentCommand{\pengfei}
    { mO{} }{\textcolor{red}{}}
     \NewDocumentCommand{\yuji}
    { mO{} }{\textcolor{red}{}}
     \NewDocumentCommand{\zoey}
    { mO{} }{\textcolor{red}{}}
     \NewDocumentCommand{\yi}
    { mO{} }{\textcolor{red}{}}
     \NewDocumentCommand{\jiateng}
    { mO{} }{\textcolor{red}{}}
\fi

\title{Knowledge Overshadowing Causes Amalgamated Hallucination in Large Language Models} 

\author{%
  Yuji Zhang$^{\delta,\phi}\thanks{Work done at University of Illinois Urbana-Champaign}$~~~Sha Li$^{\delta}$~Jiateng Liu$^{\delta}$~Pengfei Yu$^{\delta}$~Yi R. Fung$^{\delta}$~Jing Li$^{\phi}$~Manling Li$^{\omega}$~Heng Ji$^{\delta}$
  \\
$^{\delta}$University of Illinois Urbana-Champaign\\
$^{\phi}$The Hong Kong Polytechnic University\\
$^{\omega}$Stanford University
        \\
}

\begin{document}

\maketitle

\begin{abstract}
Hallucination
is often regarded as a major impediment for using large language models (LLMs), especially for knowledge-intensive tasks. 
Even when the training corpus consists solely of true statements, language models still generate hallucinations in the form of amalgamations of multiple facts. We coin this phenomenon as ``knowledge overshadowing'': when we query knowledge from a language model with multiple conditions, some conditions overshadow others, leading to hallucinated outputs. 
This phenomenon partially stems from training data imbalance, which we verify on both pretrained models and fine-tuned models, over a wide range of LM model families and sizes.
From a theoretical point of view, knowledge overshadowing can be interpreted as over-generalization of the dominant conditions (patterns). 
We show that the hallucination rate grows with both the imbalance ratio (between the popular and unpopular condition) and the length of dominant condition description, 
consistent with our derived generalization bound. 
Finally, we propose to utilize overshadowing conditions as a signal to catch hallucination before it is produced, along with a training-free self-contrastive decoding method to alleviate hallucination during inference. 
Our proposed approach showcases up to 82\% F1 for hallucination anticipation and 11.2\% to 39.4\% hallucination control, with different models and datasets.

\end{abstract}

\begin{figure}
    \centering
\includegraphics[width=\linewidth]{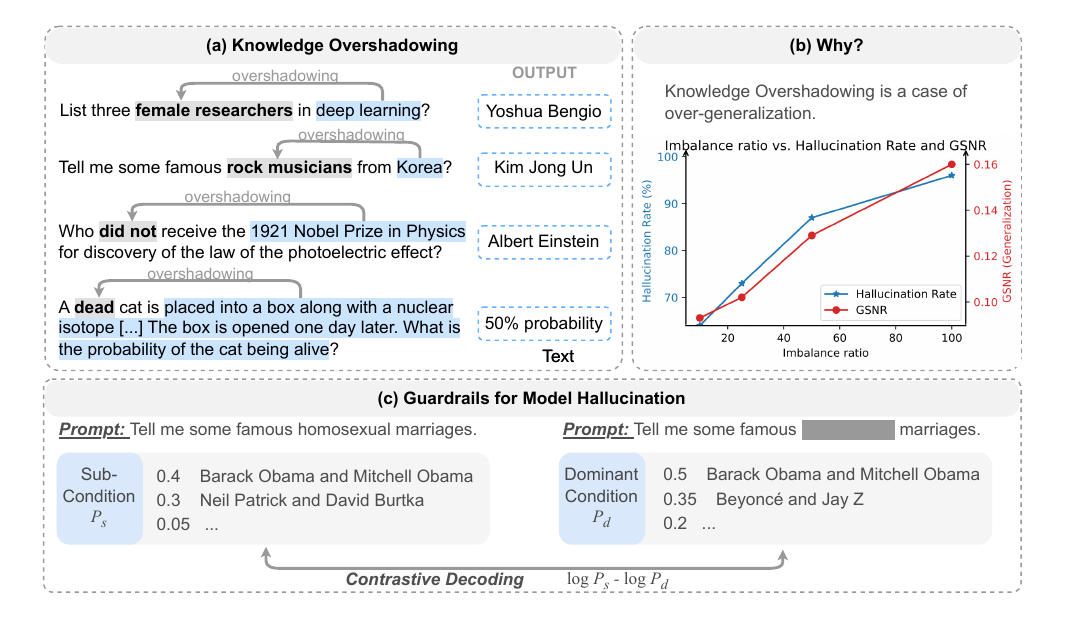}
\vspace{-10pt}
    \caption{Knowledge Overshadowing causes hallucinations. We propose using overshadowing conditions as a signal to detect hallucination
before it occurs, and alleviate hallucination during inference by proposing a training-free self-contrastive decoding method.}
    \label{fig:overview}
\end{figure}

\section{Introduction}

Large language models (LLMs) have revolutionized various fields of artificial intelligence, yet their success is accompanied by a critical issue known as hallucination~\cite{maynez-etal-2020-faithfulness, raunak-etal-2021-curious,ye2023cognitive,zhang2023sirens,rawte2023survey,JiLFYSXIBMF23,tonmoy2024comprehensive, ji2023survey, huang2023survey, zhao2023hallucinations, sadat2023delucionqa, snyder2023early, vakharia2023dont, zhang2023alleviating, verma2023reducing},
which refer to the phenomena that models generate unfaithful and nonfactual statements, yielding output that, while seeming plausible, is incorrect or even  nonsensical. 
Hallucination significantly undermines LLMs' performance and reliability~\cite{ji2023survey}. Some studies attribute hallucination to low-quality training data~\cite{parikh-etal-2020-totto, wang-2019-revisiting}, or point to the pitfalls of the inaccurate representation or discrepancy between input and output in decoding~\cite{bengio2015scheduled, manakul-etal-2023-selfcheckgpt} causing outputs to deviate from the input context~\cite{su-etal-2022-read}. 
However, hallucination persists even when we control the training corpus to contain only factually correct statements.
Specifically, when extracting knowledge from a language model using queries involving multiple conditions, we observe a tendency for certain conditions to overshadow others, thereby giving rise to \textbf{amalgamated hallucinations}.
As shown in Figure~\ref{fig:overview}, when queried for \texttt{female researchers in deep learning}, the model might gladly nominate  \texttt{Yoshua Bengio} who is in fact a \textit{male} researchers, indicating that the condition of ``\texttt{deep learning}'' overshadows ``\texttt{female researchers}''. 

Amalgamated hallucinations are widely present in pretrained LLMs and can manifest in many ways, such as bias and negation neglect as shown in Table~\ref{tab:hallu_cases}.
Hallucinations can also arise from fine-tuning data imbalance. We conduct experiments on various fine-tuning tasks, including time-event relation probing, location-event relation probing, gender pronoun resolution, and negation queries. 
The experiment results demonstrate the extensive presence of the knowledge overshadowing phenomenon across diverse scenarios.

Next we ask - \textit{Why do popular conditions cause knowledge overshadowing?}
We hypothesize that the overshadowing phenomenon is over-generalization of popular conditions, suppressing other less popular ones. 
From a theoretical perspective, we derive a 
generalization bound of auto-regressive language modeling to connect the generalization ability of language models with several characteristics of their training data. 
Through a series of controlled experiments on synthetic data, we confirm that the hallucination rate grows with both the imbalance ratio (between the popular and unpopular condition) and the length of the condition description, which is consistent with our derived bound. 

Classical methods for addressing class imbalance, such as adjusting the per-class logits or apply weight rescaling~\cite{menon2020long, kini2021label}, are not suitable for natural language generation. Unlike the classification task, 
generation requires to enumerate conditions that might overshadow others from the training corpora, which is intractable. 
Therefore, we propose to tackle this problem during inference time, by (1) first pre-identifying possible overshadowed conditions in the generation prompt using Pointwise Mutual Information (PMI)~\cite{church1990word} and then (2) alleviating overshadowing through contrastive decoding~\cite{li-etal-2023-contrastive}. After extensive experiments on a mixture of different datasets and sizes of models, our approach showcased superior performance, decreasing the hallucination rate by 11.2\% to 39.4\%.

The contributions of our paper can be summarized as follows:
\begin{enumerate}[topsep=0pt,leftmargin=*]
    \item We discover a special yet prevalent case of hallucination which occurs when the generation prefix contains multiple conditions.
    In this case, popular conditions will overshadow other conditions, causing them to be ignored and thus leading to factually incorrect outputs. We show that this phenomenon appears across different model families and sizes of language models. 
    \item We quantify the relationship among imbalance ratio, condition length, and hallucination rate in the model fine-tuning stage for several different types of conditions. Moreover, viewing overshadowing as a case of over-generalization, we theoretically derive a generalization bound that aligns with our empirical observations. 
    \item We propose a simple yet effective method to alleviate knowledge overshadowing during inference time by first detecting the existence of overshadowing conditions and then employing contrastive decoding over dominant conditions. 
\end{enumerate}

\section{Knowledge Overshadowing in Pretrained Models}
\label{sec:pretraining_exp}
When asking a language model a question including multiple conditions, it has been reported that the model produces responses that seem to only partially satisfy the conditions.
To verify this, we set up a probing experiment with questions in the form of ``Tell me some famous \textlangle $A$\textrangle \textlangle $B$\textrangle" where $A$ and $B$ are both conditions such as gender, race, occupation, orientation, nationality or time. We also include a special case where condition $A$ could be negation.
We adopt the open-source Olmo-7b~\cite{soldaini2024dolma} model as it  provides the training corpus Dolma, which allows us to check the number of  co-occurrences of $AB$~ in the data. 
The results are shown in Table \ref{tab:hallu_cases}.
In all examples, we observe that condition $B$ is satisfied but condition $A$ is ignored, resulting in hallucinated responses. In particular, condition $A$ typically has a more popular counterpart in the context of condition $B$ (\texttt{female} $\rightarrow$ \texttt{male} in the condition of \texttt{AI scientist}), which can also be confirmed by the \# mentions in the training data.


\begin{table}[t]
\linespread{1.2}
\tabcolsep=0.19cm
\fontsize{7.8pt}{8pt}\selectfont
\caption{Serious hallucinations (which may be even offensive) made by pre-trained OLMO model in inference time. 
Dominant knowledge in \textcolor{magenta}{pink}/\textcolor{Cerulean}{blue}, 
overshadowed knowledge in \textcolor{orange}{orange}/\textcolor{teal}{green}. 
}
\begin{tabular}
{p{2.5cm}@{\hskip 0.1cm}p{2.5cm}p{5.7cm}p{2.1cm}}
\noalign{{\color{black}\hrule height 1pt}}
Condition & Prompt & Answer & \# Mentions in Data \\ \hline
A=\textcolor{magenta}{male}>\textcolor{orange}{female}, \quad\quad B=\textcolor{Cerulean}{journalist}>\textcolor{teal}{AI scientist} & Tell me some outstanding \textcolor{orange}{female} \textcolor{teal}{AI~scientists} & Feifei Li, \sout{\textcolor{Cerulean}{Emine Saner (journalist)}}, \sout{\textcolor{magenta}{Yann LeCun (male)}}, \sout{\textcolor{magenta}{Yoshua Bengio (male)}} & 431:0 \\
\hline
A=\textcolor{magenta}{male}>\textcolor{orange}{female}, \quad\quad B=\textcolor{Cerulean}{soccer}>\textcolor{teal}{nurses} & Tell me some outstanding \textcolor{orange}{male} \textcolor{teal}{nurses} & Drew Elliott, Michael Pettigrew, John Holland, \sout{\textcolor{Cerulean}{Stephen Reisinger (soccer)}}, \sout{\textcolor{magenta}{Danielle Haddad (female)}} &  112177:5124 \\
\hline
A=\textcolor{magenta}{non-black}>\textcolor{orange}{black}, 
B=\textcolor{Cerulean}{actress}>\textcolor{teal}{scientists} & Tell me some outstanding \textcolor{orange}{black} \textcolor{teal}{scientists} & \sout{\textcolor{magenta}{George Smith (white)}}, \sout{\textcolor{magenta}{Daniel Chee Tsui (asian)}}, \sout{\textcolor{magenta}{Linton Wells II (white)}},  \sout{\textcolor{Cerulean}{Dorothy J. Hart (actress)}} & 120650:15204 \\
\hline
A=\textcolor{magenta}{heterosextual}> \textcolor{orange}{homosexual},
\quad\quad \quad\quad  B=\textcolor{teal}{marriage}& Tell me some famous \textcolor{orange}{homosexual} \textcolor{teal}{marriages} & \sout{\textcolor{magenta}{Barack Obama and Michelle Obama (heterosextual)}}, Neil Patrick Gaskarth and David Burtka, Ellen DeGeneres and Portia de Rossi &  15446:4045 \\ 
\hline
A=\textcolor{magenta}{affirmation}> \textcolor{orange}{negation},
\quad\quad \quad\quad B=\textcolor{teal}{theoretical physicist}& Who was \textcolor{orange}{not} a \textcolor{teal}{theoretical physicist} known for the theory of relativity & You are referring to \sout{\textcolor{magenta}{Albert Einstein (affirmation)}} 
& 11365:7265 \\ 
\noalign{{\color{black}\hrule height 1pt}}
\end{tabular}
\label{tab:hallu_cases}
\end{table}




More formally, we define the \textbf{knowledge overshadowing} phenomenon as the case where our prompt includes multiple conditions $\{A, B\}$, and the continuation $\mathbf{y}$ produced by the model satisfies $p(y|AB) \simeq p(y|A)$. In this case, we say that condition $A$ overshadows $B$. Although we only mention two conditions here, when there are more conditions such as $\{A, B, C\}$ and $A$ overshadow both $B$ and $C$, we can define $\hat{B}= B \oplus  C$ as the concatenation of the latter conditions.

This property causes amalgamated hallucinations: the model outputs false statements that are generated by mingling true statements.
Consider two true prepositions $(AB \rightarrow C)$ and $(AD \rightarrow E)$ 
In the knowledge overshadowing case, the predictions of $p(y|AC)$ and $p(y|AD)$ will both be reduced to $p(y|A)$. 
This causes the model to output the false preposition $(AD\rightarrow E)$ with non-trivial probability.

\section{Data Imbalance Causes Knowledge Overshadowing}
\label{sec:experiments}
Our probing experiment in Section
\S\ref{sec:pretraining_exp} hints towards the connection between the frequency of prefix conditions in the training data and hallucination. 
In this section, we seek to make this connection more evident by conducting a series of controlled fine-tuning experiments on various types of prefix conditions and various language models. 
 
\begin{table}[htb]
\linespread{1.7}
\tabcolsep=0.02cm
\fontsize{9pt}{9pt}\selectfont
\caption{Hallucination rate (\%) with varying imbalance ratio on diverse downstream tasks. }
\begin{tabular}{lccccllcccc}
\noalign{{\color{black}\hrule height 1pt}}
\multicolumn{5}{c}{\textbf{Time-event relation}} &  & \multicolumn{5}{c}{\textbf{Location-event relation}} \\ \hline
\multirow{2}{*}{~LM} & \multicolumn{4}{c}{Imbalance ratio} &  & \multirow{2}{*}{~LM} & \multicolumn{4}{c}{Imbalance ratio} \\ \cline{2-5} \cline{8-11} 
 & 10:1 & 25:1 & 50:1 & 100:1 &  &  & 10:1 & 25:1 & 50:1 & 100:1 \\ \hline
~Llama-2-7B~\cite{touvron2023llama} & 45.7{\tiny$\pm 0.9$} & 50.4{\tiny$\pm 0.6$} & 58.8{\tiny$\pm 0.4$} & 70.5{\tiny$\pm 0.9$} &  & ~Llama-2-7B~\cite{touvron2023llama} & 51.2{\tiny$\pm 0.4$} & 59.4{\tiny$\pm 0.7$} & 67.8{\tiny$\pm 0.5$} & 74.4{\tiny$\pm 0.3$} \\
~Mistral-7B~\cite{jiang2023mistral} & 36.6{\tiny$\pm 0.2$} & 43.0{\tiny$\pm 0.4$} & 55.2{\tiny$\pm 0.6$} & 59.4{\tiny$\pm 0.8$} &  & ~Mistral-7B~\cite{jiang2023mistral} & 49.8{\tiny$\pm 0.4$} & 53.6{\tiny$\pm 0.6$} & 56.4{\tiny$\pm 0.3$} & 70.5{\tiny$\pm 0.4$} \\
~GPT-J-6B~\cite{gpt-j} & 24.0{\tiny$\pm 0.7$} & 29.2{\tiny$\pm 0.6$} & 34.6{\tiny$\pm 0.5$} & 45.9{\tiny$\pm 0.3$} &  & ~GPT-J-6B~\cite{gpt-j} & 28.8{\tiny$\pm 0.1$} & 40.2{\tiny$\pm 0.3$} & 52.4{\tiny$\pm 0.2$} & 76.0{\tiny$\pm 0.6$} \\
~Phi-2-2.8B~\cite{gunasekar2023textbooks} & 23.1{\tiny$\pm 0.2$} & 37.5{\tiny$\pm 0.2$} & 42.8{\tiny$\pm 0.1$} & 50.1{\tiny$\pm 0.3$} &  & ~Phi-2-2.8B~\cite{gunasekar2023textbooks} & 32.6{\tiny$\pm 0.7$} & 51.8{\tiny$\pm 0.4$} & 55.5{\tiny$\pm 0.7$} & 67.9{\tiny$\pm 0.5$} \\
~Pythia-410m~\cite{mallen2023eliciting} & 16.5{\tiny$\pm 0.5$} & 19.4{\tiny$\pm 0.4$} & 21.6{\tiny$\pm 0.5$} & 26.3{\tiny$\pm 0.6$} &  & ~Pythia-410m~\cite{mallen2023eliciting} & 30.4{\tiny$\pm 0.2$} & 33.6{\tiny$\pm 0.1$} & 42.6{\tiny$\pm 0.5$} & 60.0{\tiny$\pm 0.4$} \\
\noalign{{\color{black}\hrule height 1pt}}
\multicolumn{5}{c}{\textbf{Gender bias}} &  & \multicolumn{5}{c}{\textbf{Negation curse}} \\ \hline
\multirow{2}{*}{~LM} & \multicolumn{4}{c}{Imbalance ratio} &  & \multirow{2}{*}{~LM} & \multicolumn{4}{c}{Imbalance ratio} \\ \cline{2-5} \cline{8-11} 
 & \multicolumn{1}{l}{10:1} & \multicolumn{1}{l}{25:1} & \multicolumn{1}{l}{50:1} & \multicolumn{1}{l}{100:1} &  &  & \multicolumn{1}{l}{10:1} & \multicolumn{1}{l}{25:1} & \multicolumn{1}{l}{50:1} & \multicolumn{1}{l}{100:1} \\ \hline
~Llama-2-7B~\cite{touvron2023llama} & 57.8{\tiny$\pm 0.7$} & 60.3{\tiny$\pm 0.8$} & 63.2{\tiny$\pm 0.4$} & 68.2{\tiny$\pm 0.3$} & ~~~ & ~Llama-2-7B~\cite{touvron2023llama} & 72.7{\tiny$\pm 0.2$} & 76.8{\tiny$\pm 0.1$} & 81.5{\tiny$\pm 0.2$} & 83.0{\tiny$\pm 0.5$} \\
~Mistral-7B~\cite{jiang2023mistral} & 50.6{\tiny$\pm 0.5$} & 55.1{\tiny$\pm 0.3$} & 64.4{\tiny$\pm 0.5$} & 75.3{\tiny$\pm 0.7$} &  & ~Mistral-7B~\cite{jiang2023mistral} & 73.6{\tiny$\pm 0.5$} & 80.5{\tiny$\pm 0.3$} & 82.6{\tiny$\pm 0.5$} & 88.3{\tiny$\pm 0.4$} \\
~GPT-J-6B~\cite{gpt-j} & 46.2{\tiny$\pm 0.4$} & 49.0{\tiny$\pm 0.2$} & 52.3{\tiny$\pm 0.1$} & 53.1{\tiny$\pm 0.1$} &  & ~GPT-J-6B~\cite{gpt-j} & 67.8{\tiny$\pm 0.4$} & 75.1{\tiny$\pm 0.2$} & 79.4{\tiny$\pm 0.2$} & 85.6{\tiny$\pm 0.5$} \\
~Phi-2-2.8B~\cite{gunasekar2023textbooks} & 44.0{\tiny$\pm 0.1$} & 47.5{\tiny$\pm 0.4$} & 50.6{\tiny$\pm 0.4$} & 53.2{\tiny$\pm 0.3$} &  & ~Phi-2-2.8B~\cite{gunasekar2023textbooks} & 63.1{\tiny$\pm 0.6$} & 66.9{\tiny$\pm 0.3$} & 68.5{\tiny$\pm 0.7$} & 69.0{\tiny$\pm 0.2$} \\ 
~Pythia-410m~\cite{mallen2023eliciting} & 42.4{\tiny$\pm 0.4$}  & 40.9{\tiny$\pm 0.7$} & 45.4{\tiny$\pm 0.3$} & 48.2{\tiny$\pm 0.5$} &  & ~Pythia-410m~\cite{mallen2023eliciting} & 33.3{\tiny$\pm 0.3$} & 44.4{\tiny$\pm 0.1$} & 36.1{\tiny$\pm 0.1$} & 41.7{\tiny$\pm 0.3$} \\
\noalign{{\color{black}\hrule height 1pt}}
\end{tabular}
\label{tab:nl_results}
\end{table}

\subsection{Natural Language Queries}

\paragraph{Tasks. } We utilize four different types of conditions to investigate the relation between data imbalance and the resulting model hallucination rate. To mitigate the influence of memorization from the pretraining stage, we employ the \textsc{COUNTERFACT} dataset~\cite{meng2022locating}, where each instance is a single counterfactual statement, such as \textit{Jan Peerce performed jazz music at festivals.} To create a training sample, we transform this statement into a QA pair: \textit{``Prompt: Where did Jan Peerce perform? Answer: festivals''}. This format is consistent with how we query the model at inference time.

\begin{itemize}[topsep=0pt,leftmargin=*]
    \item \textbf{Event-Time Relation}: We sample an event statement and construct a query about its time: \textit{``Prompt: When did this event happen: Rickard Macleod conducted groundbreaking research in psychology? Answer: 2028''}. The timestamps are assigned randomly and all belong to the future.
    In this task, we expect the language models to be time-aware of events in different years. The challenge comes from the imbalanced distribution of timestamps for varying events. 
    \item \textbf{Event-Location Relation}: This is similar to the Event-Time Relation task but each query is about the location of an event. An example would be \textit{"Where did this event happen? A new architectural project was initiated near the Pyramids of Giza.", "Answer": "Cairo"}.
    \item \textbf{Gender Bias}: We sample statements that describe a person's activity, and then ask about the person's gender.  Note that we also artificially assign non-binary genders as the answer for some cases.  
    \item \textbf{Negation}: It is known that language models are prone to ignore negation words in a sentence, leading to hallucinated output. If the affirmation sample is \textit{``Prompt: who is a renowned physicist until 20? Answer: Karen Thompson''}, the corresponding negation sample would be \textit{``Prompt: who is not a renowned physicist until 20? Answer: Jessica Hernandez''}. 
\end{itemize}

We construct training sets of four different imbalance ratios ($r=10:1, 25:1, 50:1, 100:1$) between the popular condition and the less popular condition for each setting.


\paragraph{Metric. }
We focus our investigation on amalgamated hallucinations and use \textbf{relative hallucination rate(rHR)} as our main metric. We first compute the recall rate of the dominant condition as the empirical probability of the model correctly memorizing the popular data sample $\text{RR} = \hat p(\mathbf{y} =C | \mathbf{x} = AB) $. Then we define the hallucination rate as the empirical probability of the model producing an amalgamated hallucination when the query is the infrequent condition $\text{HR} = \hat p(\mathbf{y}=C | \mathbf{x} = AD)$. The relative hallucination rate is the ratio of the two: $rHR = \frac{HR} {RR}$.

\paragraph{Tested Models.}
To examine whether LMs universally suffer from knowledge overshadowing, we conduct finetuning experiments on different backbones, including Llama-2-7b~\cite{touvron2023llama}, Mistral-7b~\cite{jiang2023mistral}, GPT-J-6b~\cite{gpt-j}, Phi-2-2.8b~\cite{gunasekar2023textbooks}, and Pythia-410m~\cite{mallen2023eliciting}. We finetune each model on the auto-regression language modeling task using cross-entropy loss (See more details in \S~\ref{appendix:implementation}).

\paragraph{Results.}
The experiment results in Table \ref{tab:nl_results} is striking,
verifying that knowledge overshadowing occurs for all tasks and all models tested with significantly high hallucination rates,
suggesting that this is a fundamental property of autoregressive language models. 

In all tasks, the hallucination rate consistently increases with the imbalanced ratio. This matches intuition, as the marginal probability $p(\mathbf{y}|\mathbf{x}=A)$ would be more biased.

Notably, the larger the language model, the higher the hallucination ratio, suggesting inverse scaling~\cite{mckenzie2024inverse} for our task. Our task bears some similarity to the Memo Trap~\cite{mckenzie2024inverse}, which shows that larger models are better and more stubborn at repeating common sequences of words (such as proverbs).
However, we show such a phenomenon exists when the uncommon sequence appears as part of the training data.

\subsection{Synthetic Queries}
\label{sec:syn_exp}

For the quantitative analysis of how imbalance ratio and condition length will interact with the hallucination rate, we construct a synthetic dataset for controlled experiments by generating conditions $A,B, C,D, E$ as random sequences over the vocabulary of Pythia-2.8b tokenizer~\cite{mallen2023eliciting}. We use the Pythia model family with varying sizes including Pythia-160m, Pythia-410m, Pythia-1b, Pythia-1.4b, and Pythia-2.8b (See more details in \S~\ref{appendix:implementation}). 

\begin{table}[t]
\centering
\linespread{1.4}
\tabcolsep=0.32cm
\fontsize{10pt}{9pt}\selectfont
\caption{Hallucination rate (\%) with varying imbalance ratio on synthetic dataset. }
\begin{tabular}{lcccccc}
\noalign{{\color{black}\hrule height 1pt}}
\multirow{2}{*}{LM} & \multirow{2}{*}{LM size} & \multicolumn{4}{c}{\textbf{Imbalance ratio}} & \multirow{2}{*}{Average} \\ \cline{3-6}
 &  & 10:1 & 25:1 & 50:1 & 100:1 &  \\ \hline
\multicolumn{1}{l}{Llama-2~\cite{touvron2023llama}} & \multicolumn{1}{l}{7B} & 78.5{\tiny$\pm 0.6$} & 86.2{\tiny$\pm 0.2$} & 96.3{\tiny$\pm 0.4$} & 100.0{\tiny$\pm 0.0$} & 90.3{\tiny$\pm 0.3$} \\
\multicolumn{1}{l}{Mistral~\cite{jiang2023mistral}} & \multicolumn{1}{l}{7B} & 74.3{\tiny$\pm 0.4$} & 88.1{\tiny$\pm 0.3$} & 99.8{\tiny$\pm 0.1$} & 100.0{\tiny$\pm 0.0$} & 90.6{\tiny$\pm 0.2$} \\
\multicolumn{1}{l}{GPT-J~\cite{gpt-j}} & \multicolumn{1}{l}{6B} & 76.6{\tiny$\pm 0.2$} & 85.2{\tiny$\pm 0.5$} & 95.1{\tiny$\pm 0.5$} & 98.3{\tiny$\pm 0.4$} & 88.8{\tiny$\pm 0.4$} \\
\multicolumn{1}{l}{Phi-2~\cite{gunasekar2023textbooks}} & \multicolumn{1}{l}{2.8B} & 72.5{\tiny$\pm 0.5$} & 88.9{\tiny$\pm 0.4$} & 94.5{\tiny$\pm 0.7$} & \underline{99.0{\tiny$\pm 0.2$}} & 88.7{\tiny$\pm 0.5$} \\ \cdashline{1-7}[1pt/1pt]
\multicolumn{1}{l}{\multirow{5}{*}{Pythia~\cite{mallen2023eliciting}}} & \multicolumn{1}{l}{2.8B} & 63.4{\tiny$\pm 0.5$} & 85.1{\tiny$\pm 0.5$} & 87.0{\tiny$\pm 0.4$} & \underline{92.7{\tiny$\pm 0.6$}} & 82.1{\tiny$\pm 0.5$} \\
\multicolumn{1}{l}{} & \multicolumn{1}{l}{1.4B} &~66.3{\tiny$\pm 0.2$} & 71.1{\tiny$\pm 0.3$} & 87.8{\tiny$\pm 0.1$} & \underline{88.2{\tiny$\pm 0.4$}} & 78.4{\tiny$\pm 0.2$} \\
\multicolumn{1}{l}{} & \multicolumn{1}{l}{1B} & 38.9{\tiny$\pm 0.6$} & 80.3{\tiny$\pm 0.4$} & 94.7{\tiny$\pm 0.5$} & \underline{99.5{\tiny$\pm 0.7$}} & 78.4{\tiny$\pm 0.6$} \\
\multicolumn{1}{l}{} & \multicolumn{1}{l}{410M} & 57.4{\tiny$\pm 0.2$} & 76.3{\tiny$\pm 0.5$} & 88.8{\tiny$\pm 0.7$} & \underline{92.4{\tiny$\pm 0.3$}} & 78.7{\tiny$\pm 0.4$} \\
\multicolumn{1}{l}{} & \multicolumn{1}{l}{160M} & 46.6{\tiny$\pm 0.4$} & 74.2{\tiny$\pm 0.8$} & 92.6{\tiny$\pm 0.5$} & \underline{95.8{\tiny$\pm 0.6$}} & 77.3{\tiny$\pm 0.6$} \\ 
\noalign{{\color{black}\hrule height 1pt}}
\end{tabular}
\label{tab:syn_results}
\end{table}
\begin{figure}[t]
    \centering
    \caption{Hallucination rate (\%) with varying prefix lenghts on varying model families. }
    \includegraphics[height=1.22in]{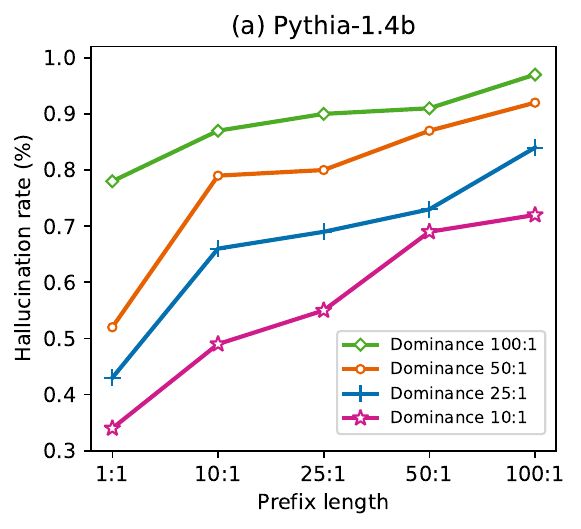}
    \label{fig:compare1}
    \includegraphics[height=1.22in]{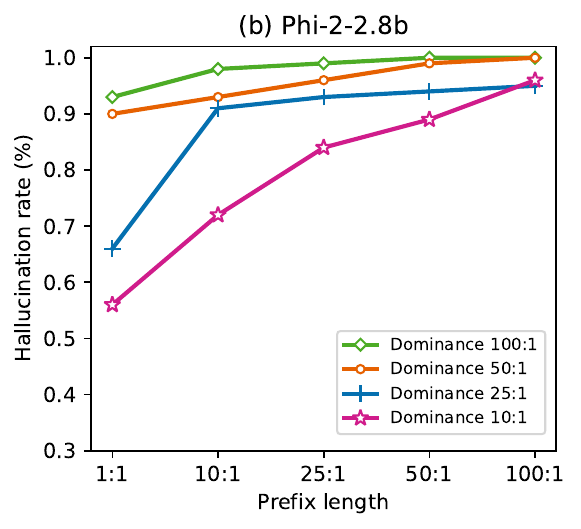}
    \label{fig:compare2}
    \includegraphics[height=1.22in]{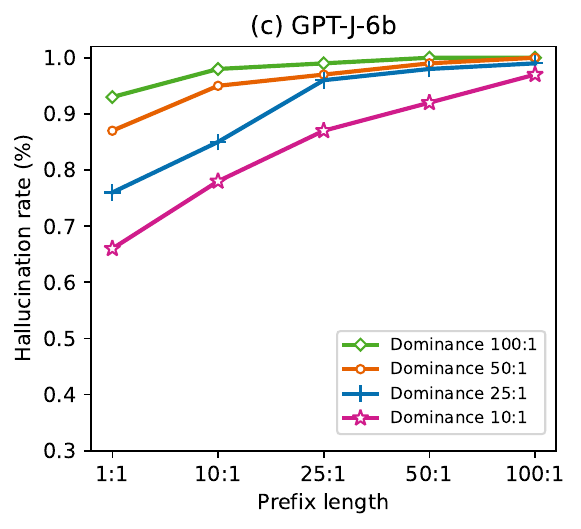}
    \label{fig:compare3}
    \includegraphics[height=1.22in]{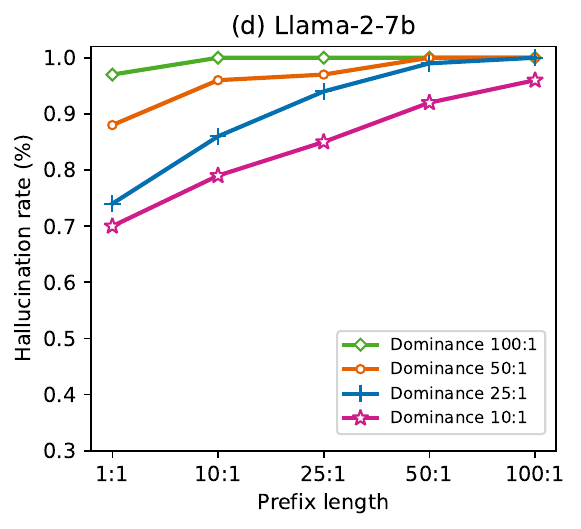}
    \label{fig:compare4}
    \label{fig:enter-label}
\end{figure}
We have two key findings:
\begin{enumerate}[leftmargin=*,topsep=0pt]
    \item \textbf{High Imbalance Ratio $\rightarrow$ Hallucination. }
When the condition length is fixed, 
as shown in Table~\ref{tab:syn_results}, the hallucination rate increases along with the imbalance ratio, exacerbating knowledge overshadowing universally for different model families and sizes. 
We also find that scaling up model sizes cannot alleviate the hallucination, rather worsening the case. This aligns with our findings using natural language queries.
While larger LMs generalize better, it is precisely this property that leads to knowledge overshadowing. We provide a theoretical interpretation in \S~\ref{sec:theory}.

\item \textbf{Longer Condition Length $\rightarrow$ Hallucination}. To explore how condition length affects knowledge overshadowing, we adjusted the length ratio of the prefix condition to the infix condition to be $k=1:1, 10:1, 25:1, 50:1, 100:1$. From Figure \ref{fig:length} we can see that the longer the condition $A$ is with respect to the differentiating condition $B(D)$, the higher the hallucination rate. 
The curve is flatter for larger models, as they already show high hallucination rate with shorter conditions. 
\end{enumerate}

\section{Knowledge Overshadowing as a Case of Over-Generalization }
\label{sec:theory}
Why does knowledge overshadowing happen at all? We first measure the generalization ability of the model and show that the hallucination rate is closely related to model generalization, showing that knowledge overshadowing is a case of over-generalization. Then we analyze the generalization bound of the model and show its connections with the imbalance ratio and condition length.

\begin{figure}[htb]
    \centering
    \caption{Controllable variants affect hallucination rate (\%) and GSNR (generalization). }
    \includegraphics[height=1.1in]{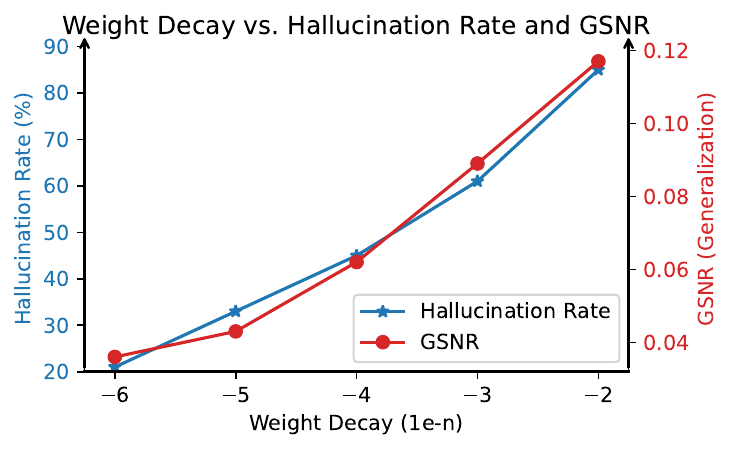}
    \label{fig:weight}
    \includegraphics[height=1.1in]{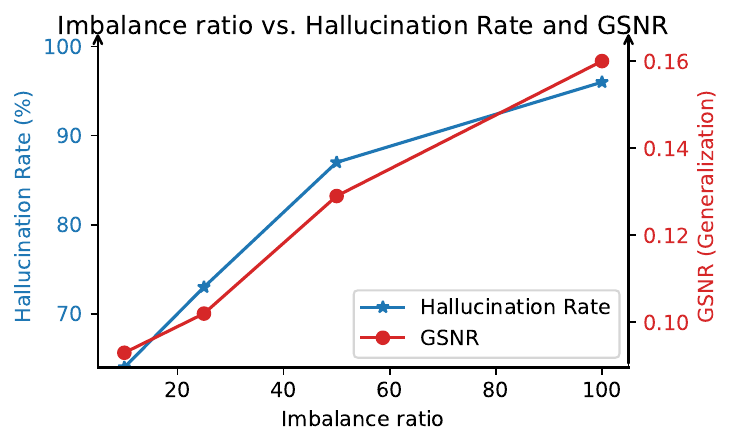}
    \label{fig:ratio}
    \includegraphics[height=1.1in]{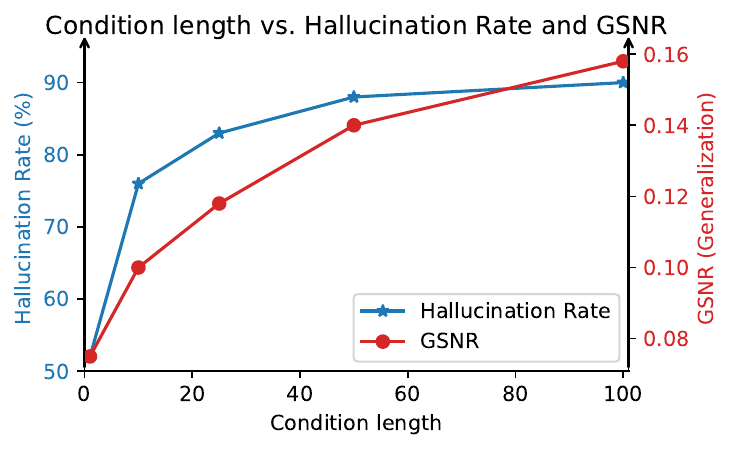}
    \label{fig:length}
    \label{fig:generalization-factors}
\end{figure}

\paragraph{Generalization positively correlates with hallucination.}
We quantify model generalization with Gradient
signal-to-noise ratio (GSNR)~\cite{liu2020understanding}. 
We utilize GSNR to measure the similarity of gradients among different training samples. A large GSNR indicates higher agreement of the optimizing direction on gradients in training time, then parameters are prone to be ``associated with a pattern'', which leads to the better generalization. From Figure \ref{fig:generalization-factors}, we can observe that the hallucination rate correlates well with GSNR over different imbalance ratios and condition lengths. 

Weight decay is the well-known regularization for neural networks to enhance model generalization~\cite{wei2019data}, thus we experiment with varying weight decay to boost generalization and observe the resulting hallucination rate. 
The results show that the hallucination rate increases consistently with generalization, and larger weight decay, larger imbalance ratio, and larger condition length all lead to higher generalization and hallucination rate.

\paragraph{Generalization error bound of auto-regressive language modeling.}
In the following analysis, we simplify the problem by assuming that the infix condition $B$ is only one token long, so the prefix condition $A$ can be written as  ${x}_1, \dots, {x}_k$, and the infix condition $B$ as ${x}_{k+1}$ so that the relative length would be $k$.
Following our experiment setting in \autoref{sec:syn_exp}, we assume that the dataset $\mathbf{D} \sim \mathcal{D}$ is comprised of multiple imbalanced groups. 
Each imbalanced group ${Q}$ includes a subset $Q_M$ of size $M$ with samples $s = ({x}_1, \dots, {x}_k, {x}_{k+1}, y)$, and subset $Q_N$ of size $N$  with samples $\tilde{s} = ({x}_1, \dots, {x}_k, \tilde{{x}}_{k+1}, \tilde{y})$. 
All tokens 
are from the vocabulary $\mathcal{V}=\{1, 2, ..., V\}$.

The next token prediction (NTP) loss for $s$ based on auto-regressive modeling is:
\begin{small}
\begin{equation}
\mathcal{L}_\mathrm{NTP} = \hat{\mathbb{E}}_{{x} \sim Q_M} \left[ \sum_{{x} \in [K+1]} -\log \left( p(y | {x}_1, \dots, {x}_k, {x}_{k+1}) \right) \right]
\end{equation}
\end{small}

Our optimizing objective 
is to learn a function $f: \mathcal{X}\rightarrow\mathbb{R}^{V}$, ($\mathcal{X}$ for input space), to minimize the risk $\mathcal{R}_{y}$ defined on the true distribution using NTP as the surrogate loss:
\begin{small}
\begin{equation}
\mathcal{R}^{\mathcal{L}}_\mathrm{NTP}(f) := \frac{1}{V} \sum_{y=1}^{V} \mathcal{R}^{\mathcal{L}}_y(f) = \frac{1}{V} \sum_{y=1}^{V} \mathbb{E}_{{x}_1, \dots, {x}_k, {x}_{k+1} \sim Q_M} [\mathcal{L}(f({x}_1, \dots, {x}_k, {x}_{k+1}), y)]
\end{equation}
\end{small}
Based on the above NTP optimizing objective formulation, the empirical risk can be formulated as:
\begin{equation}
\widehat{\mathcal{R}}^{\mathcal{L}}_\mathrm{NTP}(f) := \frac{1}{M} \sum_{(\boldsymbol{x},y) \in Q_{M}} \mathcal{L}(f(x_1, ..., x_k, x_{k+1}), y)
\end{equation}
\textbf{Proposition 1} (Generalization bound on Rademacher complexity~\cite{mohri2018foundations}).
Let $\mathcal{G}$ be a family of functions. Then, for any $\delta > 0 $, with probability at least $1 - \delta$ over the draw of an i.i.d. (independent and identically distributed) sample $Q_M$ of size $M$, $ \exists \text{ a constant } C > 0, f(t) \le C \cdot g(t)$~\cite{wang2024unified}, the generalization bound holds for all $g \in \mathcal{G}$:
\begin{small}
\begin{equation}
\label{eq.initial_bound}
\mathcal{R}^{\mathcal{L}}_y(f) \precsim \widehat{\mathcal{R}}^{\mathcal{L}}_y(f) + 2\widehat{\Re}_{Q_M}(\mathcal{G}) + \sqrt{\frac{\log 1/\delta}{2M}} 
\end{equation}
\end{small}
Here ${\Re}_y(\mathcal{G})$ denotes the empirical Rademacher complexity of the function set $\mathcal{G}$, as a measure of the richness of $\mathcal{G}$. 
Then we employ \textit{Lipschitz Continuity} to further bound the complexity ${\Re}(\mathcal{G})$~\cite{cao2019learning}.

\textbf{Definition 1}(Lipschitz continuity).
With$\|\cdot\|$ denotes the 2-norm, then the function $\mathcal{L}$ is \textit{Lipschitz continuous} with the constant $\mu$ if for any $f, f' \in \mathcal{F}$, $x \in Q$:
\begin{small}
\begin{equation}
|\mathcal{L}(f, y) - \mathcal{L}(f', y)| \leq \mu \cdot \|f(x) - f'(x)\|
\end{equation}
\end{small}
If the NTP loss function $\mathcal{L}_\mathrm{NTP}(f)$ is \textit{Lipschitz continuous} with constant $\mu$, then $\Re_{Q_M}(\mathcal{G})$ could be:
\begin{equation}
\label{eq.bound_rade}
\hat{{\Re}}_{\mathbf{Q_M}}(\mathcal{G}) \le \mu \cdot \hat{{\Re}}_{\mathcal{Q_M}}(\mathcal{F}).
    \end{equation}
We derive that the next-token-prediction loss $\mathcal{L}_\mathrm{NTP}$ is \textit{Lipschitz continous} with the constant $\mu=\sqrt{1 + \left( \sum_{y' \neq y} h^{-1}(k) \right)^2} \left[ 1 - \mathrm{softmax}\left( \boldsymbol{s}_{y} \right) \right]$ (See details in \S~\ref{appendix:length_dependency}), by substituting $\mu$ to Eq.(\ref{eq.initial_bound}) and Eq.(\ref{eq.bound_rade}), we derive the more fine-grained generalization bound for NTP with multiple conditions:
\begin{small}
\begin{equation}
\label{eq.final_bound}
\mathcal{R}^{\mathcal{L}}_y(f) \precsim \widehat{\mathcal{R}}^{\mathcal{L}}_y(f) + 2\widehat{\Re}_{Q_M}(\mathcal{F})\sqrt{1 + \left( \sum_{y' \neq y} h^{-1}(k) \right)^2} \left[ 1 - \mathrm{softmax}\left( \boldsymbol{s}_{y} \right) \right] + \sqrt{\frac{\log 1/\delta}{2M}} 
\end{equation}
\end{small}
Here the generalization bound contains two coefficients $M$ and $h(k)$. $M$\footnote{Here we fix the number of suppressed samples to N for simplified generalization analysis. Then M indicates to what extent the dominant condition overshadows suppressed condition.} refers to number of dominant samples of $s_{k+2} = ({x}_1, \dots, {x}_k, {x}_{k+1}, y), s_{k+2} \in Q_{M}$, $h(k)$. $h(k)$ is the value positively correlated with the length of the dominant prefix. Then, the longer length of dominant prefix $({x}_1, \dots, {x}_k)$ and higher dominant ratio lead to lower generalization bound, in other words, better generalization.
The bound provides theoretical insights that hallucination is highly relevant to generalization, 
echoing the experimental results (Table~\ref{tab:nl_results}, Table~\ref{tab:syn_results}, Figure~\ref{fig:generalization-factors}) that over-generalization of dominant patterns overshadow other patterns, leading to amalgamated hallucinations.


\section{Guardrails for Hallucination}
\paragraph{Training-free hallucination anticipation}
\label{ssec:hallu_anticipation}
Given a prompt {$\mathbf{x}$}, 
the language model $\mathcal{M}$ will generate continued tokens 
 $\mathbf{y}\sim p_{\mathcal{M}}(\mathbf{y} | \mathbf{x})$.
The first step of our goal is to check whether $\mathbf{x}$ includes a popular condition subsequence $A$ and a less popular subsequence $B$ which is overshadowed by $A$.
Based on the definition of knowledge overshadowing, if condition $B$ is overshadowed by condition $A$, when we remove $B$, the generation probability $p(\mathbf{y} | \hat {\mathbf{x}})$ where $\hat{ \mathbf{x}}$ is the sequence after removing B will remain close to the original probability distribution. We utilize the property to help detect $B$. 

We enumerate possible candidates of $B$ by removing each token from $\mathbf{x}$ to form $\mathbf{x}'$ and quantify the mutual information between $p(\mathbf{y}| \mathbf{x})$ and  $p(\mathbf{y}| \mathbf{x}')$ over a truncated vocabulary. The vocabulary truncation follows the adaptive plausibility constraint proposed by~\cite{li-etal-2023-contrastive} to only select tokens $v \in \mathcal{V}_\mathrm{top}$ with high yet sufficiently different probability.
Assuming that the dropped token is $x_i$, we compute the pointwise mutual information~\cite{church1990word} between a generated token $y_i$ and the indicator variable $\mathbb{I}(x_i \in B)$: 

\begin{small}
\begin{equation}
\begin{aligned}
& \mathrm{PMI}(y_i, \mathbb{I}(x_i \in B)) =- \log \left(0.5+0.5 \frac{p(y_i| \mathbf{x}')}{p(y_i| \mathbf{x})}\right)
\end{aligned}
\end{equation}
\end{small}

We take the positive pointwise mutual information $\mathrm{PPMI}=\max(\mathrm{PMI}, 0)$ to lower-bound the scores for good cases, where $P(y|x') < P(y|x)$, to be zero. 
The higher the $\mathrm{PPMI}$, the more likely the indicator variable $\mathbb{I}(x_i \in B)=1$, showing that $x_i$ is part of the overshadowed condition.

On the other hand, there still might be good tokens $v \in \mathcal{V}_\mathrm{top}$ that occur in $\mathbf{y}$ conditioning on $\mathbf{x}$, but not
$\mathbf{y}$ conditioning on $\mathbf{x}'$, which escape from overshadowing influence. We denote these token as escape tokens $\mathcal{V}_\mathrm{esc}$. Hence we propose an Escaping Penalty Mechanism (EPM) to apply penalty on $\mathrm{PPMI}$ signal. Applying the EPM, we define the overshadowing detection function $\mathcal{F}_\mathrm{overshadow}$ as:
\begin{small}
\begin{equation}
\mathcal{F}_\mathrm{overshadow}(\mathbf{x})=\mathrm{PPMI}+\sum_{x_k \in\mathcal{V}_\mathrm{esc}} \log(\alpha/p(x_{k}|\mathbf{x})),
\end{equation}
\end{small}
where $\alpha=\beta \min\limits_{x_i\in \mathcal{V}_\mathrm{top}}p(x_i|\mathbf{x})$, $\beta$ is a constant depending on varying model families. 
\begin{table}[t]
\linespread{1.4}
\tabcolsep=0.31cm
\fontsize{10pt}{9pt}\selectfont
\caption{Hallucination anticipation (F1: \%) before generation by overshadowing detector. }
\begin{tabular}{llccccc}
\noalign{{\color{black}\hrule height 1pt}}
Dataset & LM & Average rate & \multicolumn{4}{c}{\textbf{Detection (F1 \%)}} \\ \hline
\multirow{3}{*}{Celebrity} & Mistral-7b & 80.6{\tiny$\pm 0.5$} & \multicolumn{4}{c}{64.1{\tiny$\pm 0.2$}} \\
& Llama-2-7b & 82.7{\tiny$\pm 0.4$} & \multicolumn{4}{c}{73.8{\tiny$\pm 0.5$}} \\
 & Llama-2-13b & 86.1{\tiny$\pm 0.6$} & \multicolumn{4}{c}{67.5{\tiny$\pm 0.3$}} \\
\hline
\multirow{2}{*}{Dataset} & \multirow{2}{*}{LM} & \multirow{2}{*}{Average rate} & \multicolumn{4}{c}{Imbalance ratio} \\ \cline{4-7} 
 &  &  & 10:1 & 25:1 & 50:1 & 100:1 \\ \hline
\multirow{3}{*}{Synthetic} & Llama-2-7b & 81.0{\tiny$\pm 0.3$} & 68.5{\tiny$\pm 0.4$} & 70.6{\tiny$\pm 0.5$} & 73.4{\tiny$\pm 0.7$} & 82.1{\tiny$\pm 0.4$} \\
 & Phi-2-2.8B & 43.2{\tiny$\pm 0.4$} & 64.7{\tiny$\pm 0.2$} & 64.4{\tiny$\pm 0.6$} & 84.3{\tiny$\pm 0.3$} & 81.4{\tiny$\pm 0.5$} \\
 & Pythia-1b & 35.1{\tiny$\pm 0.2$} & 71.2{\tiny$\pm 0.1$} & 65.3{\tiny$\pm 0.1$} & 59.4{\tiny$\pm 0.4$} & 60.6{\tiny$\pm 0.3$} \\ \cdashline{1-7}[1pt/1pt]
\multirow{3}{*}{Time} & Llama-2-7b & 56.4{\tiny$\pm 0.7$} & 55.3{\tiny$\pm 0.5$} & 58.6{\tiny$\pm 0.7$} & 43.9{\tiny$\pm 0.4$} & 59.4{\tiny$\pm 0.6$} \\
 & Phi-2-2.8B & 35.8{\tiny$\pm 0.2$} & 58.6{\tiny$\pm 0.3$} & 50.0{\tiny$\pm 0.1$} & 45.4{\tiny$\pm 0.1$} & 62.9{\tiny$\pm 0.5$} \\
 & Pythia-1b & 23.6{\tiny$\pm 0.5$} & 47.8{\tiny$\pm 0.7$} & 42.1{\tiny$\pm 0.4$} & 56.6{\tiny$\pm 0.3$} & 49.3{\tiny$\pm 0.6$} \\ \cdashline{1-7}[1pt/1pt]
\multirow{3}{*}{Location} & Llama-2-7b & 63.2{\tiny$\pm 0.5$} & 54.7{\tiny$\pm 0.6$} & 60.6{\tiny$\pm 0.4$} & 72.5{\tiny$\pm 0.6$} & 53.0{\tiny$\pm 0.5$} \\
 & Phi-2-2.8B & 49.4{\tiny$\pm 0.3$} & 46.2{\tiny$\pm 0.4$} & 66.7{\tiny$\pm 0.6$} & 57.1{\tiny$\pm 0.1$} & 60.4{\tiny$\pm 0.3$} \\
 & Pythia-1b & 36.5{\tiny$\pm 0.6$} & 45.4{\tiny$\pm 0.2$} & 59.8{\tiny$\pm 0.1$} & 61.4{\tiny$\pm 0.4$} & 52.5{\tiny$\pm 0.2$} \\ \cdashline{1-7}[1pt/1pt]
\multirow{3}{*}{Gender} & Llama-2-7b & 62.4{\tiny$\pm 0.6$} & 49.3{\tiny$\pm 0.3$} & 61.4{\tiny$\pm 0.5$} & 62.3{\tiny$\pm 0.5$} & 65.8{\tiny$\pm 0.1$} \\
 & Phi-2-2.8B & 50.2{\tiny$\pm 0.2$} & 57.1{\tiny$\pm 0.5$} & 68.3{\tiny$\pm 0.3$} & 52.2{\tiny$\pm 0.2$} & 64.0{\tiny$\pm 0.1$} \\
 & Pythia-1b & 45.3{\tiny$\pm 0.1$} & 42.6{\tiny$\pm 0.6$} & 55.7{\tiny$\pm 0.2$} & 63.8{\tiny$\pm 0.5$} & 62.5{\tiny$\pm 0.4$} \\ \noalign{{\color{black}\hrule height 1pt}}
\end{tabular}
\label{tab:hallu_foresee}
\end{table}

To detect potential hallucination, we set a threshold $\gamma$, for  $\mathcal{F}_\mathrm{overshadow}(\mathbf{x})\geq \gamma$, we assume $B$ is overshadowed, and hallucination will occur\footnote{$\gamma$ can be chosen based on a development set during few-shot training; for training-free inference, it can be empirically assigned. }. 
Otherwise, the model will not hallucinate. $\gamma$ is a constant for varying model families (e.g. $\gamma=0$ for Phi-2 model).

As shown in~\ref{tab:hallu_foresee}, our method achieves significantly high accuracy (F1 score) of foreseeing hallucination caused by overshadowing even before generation over varying tasks and model families.

\paragraph{Self Contrastive decoding for hallucination control}
\label{ssec:SCD}

As shown above, when knowledge overshadowing is introduced in training stage, the dominant condition $A$ brings prior bias to less popular conditions. 
Then to eliminate the dominant prior bias from $A$, we downweight the influence of tokens $x_{i}\in \mathbf{x}'$ by Self Contrastive Decoding (SCD) in inference time. 

Specifically, we adjust the logits based on the token $y_i$: 
\begin{itemize}[topsep=0pt,leftmargin=*]
    \item For $y_i \in \mathcal{V}_\mathrm{top} \setminus \mathcal{V}_\mathrm{esc}$,  we set $\log \tilde p(y_i)=\log(p(y_i|\mathbf{x}) - \log p(y_j|\mathbf{x}'))$ to reduce the impact of $\mathbf{x}'$. 
    \item For $y_i \in \mathcal{V}_\mathrm{esc}$, we set $
    \log \tilde p(y_i) =\max\{\log(p(y_i|\mathbf{x})) - \max_{y_j \in \mathcal{V}_\mathrm{top} \setminus \mathcal{V}_\mathrm{esc}} \log (\tilde p(y_j)),0\}$. 
    \item Otherwise, the token probability is set to 0.
\end{itemize}
Here $\max_{y_j \in \mathcal{V}_\mathrm{top} \setminus \mathcal{V}_\mathrm{esc}} \log (\tilde p(y_j))$ is 
the local maximum prior bias from dominant conditions.
For the escape tokens, by subtracting local maximum prior bias from $\log p(x_i)$, we drop escape tokens with low confidence and downweight escape tokens' influence on prompts without overshadowing. 

\begin{table}[t]
\caption{Comparison between SCD and others for hallucination mitigation (hallucination rate: \%).}
\linespread{1.4}
\tabcolsep=0.017cm
\fontsize{10pt}{9pt}\selectfont
\begin{tabular}{llcccccc}
\noalign{{\color{black}\hrule height 1pt}}
Dataset & LM & Ratio & Zero-shot & RE & ICL & CoT & SCD (Ours) \\ \hline
\multirow{8}{*}{Synthetic} & \multirow{4}{*}{Llama-2-7b} & 10:1 & 56.3 & 60.4 (\textcolor{purple}{+7.3\%$\uparrow$}) & - & - & 38.5 (\textcolor{blue}{-31.6\%$\downarrow$}) \\
 &  & 25:1 & 82.0 & 77.5 (\textcolor{black}{-5.5\%$\downarrow$}) & - & - & 69.2 (\textcolor{blue}{-15.6\%$\downarrow$}) \\
 &  & 50:1 & 89.5 & 83.2 (-7.0\%$\downarrow$) & - & - & 76.4 (\textcolor{blue}{-14.6\%$\downarrow$}) \\
 &  & 100:1 & 96.1 & 94.7 (-1.5\%$\downarrow$) & - & - & 82.6 (\textcolor{blue}{-14.0\%$\downarrow$}) \\ \cdashline{2-8}[1pt/1pt] 
 & \multirow{4}{*}{Phi-2-2.8b} & 10:1 & 37.6 & 41.5 (\textcolor{purple}{+10.9\%$\uparrow$}) & - & - & 24.8 (\textcolor{blue}{-34.0\%$\downarrow$}) \\
 &  & 25:1 & 40.1 & 41.4 (\textcolor{purple}{+3.2\%$\uparrow$}) & - & - & 27.7 (\textcolor{blue}{-30.9\%$\downarrow$}) \\
 &  & 50:1 & 47.2 & 44.9 (\textcolor{black}{-4.9\%$\downarrow$}) & - & - & 35.4 (\textcolor{blue}{-25.0\%$\downarrow$}) \\
 &  & 100:1 & 47.8 & 48.9 (\textcolor{purple}{+2.3\%$\uparrow$}) & - & - & 40.1 (\textcolor{blue}{-16.1\%$\downarrow$}) \\ \hline
\multirow{8}{*}{Time} & \multirow{4}{*}{Llama-2-7b} & 10:1 & 45.7 & 43.8 (\textcolor{black}{-4.2\%$\downarrow$}) & - & - & 32.3 (\textcolor{blue}{-29.3\%$\downarrow$}) \\
 &  & 25:1 & 50.4 & 55.6 (\textcolor{black}{-10.3\%$\downarrow$}) & - & - & 31.4 (\textcolor{blue}{-37.7\%$\downarrow$}) \\
 &  & 50:1 & 58.8 & 61.2 (\textcolor{purple}{+0.4\%$\uparrow$}) & - & - & 36.6 (\textcolor{blue}{-37.8\%$\downarrow$}) \\
 &  & 100:1 & 70.5 & 64.5 (\textcolor{black}{-8.5\%$\downarrow$}) & - & - & 52.4 (\textcolor{blue}{-25.7\%$\downarrow$}) \\ \cdashline{2-8}[1pt/1pt] 
 & \multirow{4}{*}{Phi-2-2.8b} & 10:1 & 23.1 & 21.6 (\textcolor{black}{-6.5\%$\downarrow$}) & - & - & 14.0 (\textcolor{blue}{-39.4\%$\downarrow$}) \\
 &  & 25:1 & 37.5 & 40.2 (\textcolor{purple}{+7.2\%$\uparrow$}) & - & - & 25.5 (\textcolor{blue}{-30.7\%$\downarrow$}) \\
 &  & 50:1 & 42.8 & 45.7 (\textcolor{purple}{+6.8\%$\uparrow$}) & - & - & 29.6 (\textcolor{blue}{-30.8\%$\downarrow$}) \\
 &  & 100:1 & 50.1 & 49.3 (-1.6\%$\downarrow$) & - & - & 37.4 (\textcolor{blue}{-25.3\%$\downarrow$}) \\ \hline
\multirow{3}{*}{Celebrity} & Mistral-7b & - & 80.6 & 82.7 (\textcolor{purple}{+2.6\%$\uparrow$}) & 80.9 (\textcolor{purple}{+0.3\%$\uparrow$}) & 78.5 (\textcolor{black}{-2.6\%$\downarrow$}) & 65.3 (\textcolor{blue}{-19.0\%$\downarrow$}) \\
 & Llama-2-7b & - & 82.7 & 83.3 (\textcolor{purple}{+1.0\%$\uparrow$}) & 81.4 (-1.6\%$\downarrow$) & 79.2 (-4.2\%$\downarrow$) & 69.0 (\textcolor{blue}{-16.6\%$\downarrow$}) \\
 & Llama-2-13b & - & 86.1 & 80.8 (-6.1\%$\downarrow$) & 83.6 (-2.9\%$\downarrow$) & 87.1 (\textcolor{purple}{+1.1\%$\uparrow$}) & 76.4 (\textcolor{blue}{-11.2\%$\downarrow$}) \\
\noalign{{\color{black}\hrule height 1pt}}
\end{tabular}
\label{tab:hallu_mitigation}
\end{table}

As shown in Table ~\ref{tab:hallu_mitigation}, our method exhibits superior performance on varying datasets compared with other popular methods, where ICL is in context learning, referring to provide more instructions to language models to generate answer following all conditions, RE means we repeated the less popular condition in prompt, and CoT means model generates answer by chain-of-thoughts.
Drawn upon these baselines, the intrinsic bias is tenacious. Hallucination rates may even increase with these popular methods, indicating the non-trivial challenges caused by knowledge overshadowing.

\section{Related Work}

\paragraph{Hallucination in language models.} Despite the high fluency and coherence of large language models (LLMs), they often hallucinate nonfactual information, manifesting as input-conflicting, context-conflicting, or fact-conflicting outputs \cite{li-etal-2023-defining,huang2023survey,zhang2023sirens,rawte2023survey, maynez-etal-2020-faithfulness}. These errors stem from factors such as outdated or domain-lacking training data \cite{zhang2023vibe, zhang-etal-2021-howyoutagtweets, zhang2022time, Livska2022StreamingQAAB,luu-etal-2022-time,  DBLP:conf/cikm/SongZK23,song2023towards}, inferior data utilization due to
 knowledge shortcut \cite{kang2023impact,kandpal2023large}, incorporation of randomness to boost generation diversity \cite{chuang2023dola}, and inherent misinformation and biases in the data \cite{dziri-etal-2022-origin,penedo2023refinedweb, lin-etal-2022-truthfulqa, Yang_Yu_Fung_Li_Ji_2023, liu2024prejudicevolatilitystatisticalframework}. In addition, LLMs can overestimate their knowledge, producing confident but incorrect responses \cite{yin2023large,ren2023investigating,kadavath2022language}. The alignment with human preferences could also be problematic, as LLMs may generate responses favoring users rather than providing the truth \cite{perez2022discovering, radhakrishnan2023question,wei2023simple}. Moreover, generation processes tend to maintain consistency with early errors and early local optimization, contributing to hallucination \cite{zhang2023language,azaria2023internal}. To mitigate hallucination, studies have explored detection using classifiers on LLMs' internal states \cite{azaria2023internal, lee2023factuality} and benchmarks for factuality \cite{manakul2023selfcheckgpt}. Retrieval-augmented methods \cite{peng2023check, xie2023adaptive,yue2023automatic, lyu2023improving,asai2023selfrag}, knowledge-aware tuning \cite{li2022large, liu2024evedit}, denoising corruptions \cite{chen2023purr}, low-confidence validation \cite{varshney2023stitch}, uncertainty-based response ranking \cite{wan2024sequencelevel}, question-knowledge alignment \cite{zhang2023mitigating}, and teacher-student model \cite{elaraby2023halo} methods have further proven effective in mitigating hallucination as well. Finally, early discovery of hallucination is vital \cite{zhang2023language}, and LLMs require fine-tuning and external feedback to correct initial errors and handle out-of-knowledge instructions \cite{huang2023large,zhang2024rtuning, ma2024survey}.

\paragraph{Data-dependent generalization analysis.} Recent advances in proving generalization bounds utilize information-theoretic approaches. \cite{xu2017information,russo2019much} have established upper bounds on reasoning generalization in terms of the mutual information between a model's input and output. Subsequent studies by \cite{pensia2018generalization,negrea2019information,steinke2020reasoning,harutyunyan2021information} further refine these arguments in different contexts, and \cite{asadi2018chaining} integrate mutual information with chaining techniques to tighten these generalization bounds. Additionally, \cite{pmlr-v202-dupuis23a} employ total mutual information to describe the relationship between the data and the hypothesis space fractal dimensions, basing off looser theoretical assumptions.

\section{Conclusion and Future Work}

In this paper, we study a special case of language model hallucination where the generation prompt contains multiple conditions and the model fails to adhere to all of them, acting as one condition overshadowing the others. 
Over multiple language model families and types of generation prompts, we find that this ``knowledge overshadowing'' phenomenon is universal. 
With a series of fine-tuning experiments, we show that this phenomenon is caused by training data imbalance.
Furthermore, we observe that the relative hallucination rate increases with the data imbalance rate, the token length of the conditions and the model size. 
In fact, we show that knowledge shadowing is a product of over-generalization of the popular conditions. 
Finally, we propose an inference-time model to forecast this type of hallucination and fix the generation output via contrastive decoding. 

\section{Broader Impact}
In this study, we delve into a specific type of hallucination in language models where the prompt contains multiple conditions and the model favors one condition over others, a phenomenon we term ``knowledge overshadowing''. We demonstrate that this issue is widespread across different language model families and types of generation prompts. Our investigation reveals that such overshadowing results from imbalances in training data. Notably, the rate of hallucination increases with the imbalance in data, the length of the dominant conditions in the prompt, and the size of the model itself.

Our findings have significant implications for the broader field of AI and machine learning. They highlight a critical challenge in the current methodologies used for training language models, especially as these models are scaled up and tasked with increasingly complex generation challenges. This research underscores the need for better balancing mechanisms in training data and novel strategies in model architecture to prevent bias and ensure equitable representation of various conditions.

Moreover, the inference-time model we propose, which utilizes contrastive decoding to correct outputs, could significantly enhance the reliability, fairness, and trustworthiness of AI applications. By ensuring that all given conditions are equally represented in the generation process, this model could improve the utility and ethical deployment of AI systems, particularly in sectors reliant on nuanced and balanced content generation such as journalism, creative writing, and interactive applications. Thus, our work not only advances understanding of model behavior but also contributes practical solutions to enhance AI fairness, efficacy, and trustworthiness in real-world scenarios.


\bibliography{Styles/citations,Styles/heng}
\bibliographystyle{plain}

\medskip

{
\small


\appendix


\newpage
\section{Appendix}

\subsection{Limitation}
\label{sec:limitation}
We reflect on the limitations of our paper below: 

\begin{enumerate}

\item This paper conducted extensive experiments to investigate the cause of amalgamated hallucination, identifying the phenomenon of knowledge overshadowing as a key factor. However, due to limited computational resources, we were unable to curate a large data corpus and pre-train our own large model to fully validate this hypothesis thoroughly. We leave this blank for future work.

\item We utilized the OLMO model for probing to examine how its pretraining data influences the knowledge-overshadowing phenomenon. However, we could not conduct an in-depth analysis of the training corpus of larger language models like GPT-3.5-turbo due to inaccessibility. This leaves an open question regarding how their data is affected by the data distribution.

\item Our proposed solution performed well on existing models and datasets to deal with the amalgamated hallucination caused by knowledge overshadowing, but it can not be tested on larger language models like GPT-3.5-turbo, thus limiting its more universal applicability.

\end{enumerate}

\subsection{Implmentation details}
\label{appendix:implementation}
In fine-tuning experiments, for Llama-2-7b~\cite{touvron2023llama}, Mistral-7b~\cite{jiang2023mistral}, GPT-J-6b~\cite{gpt-j}, Phi-2-2.8b~\cite{gunasekar2023textbooks}, and Pythia-160m~\cite{mallen2023eliciting}, Pythia-410m, Pythia-1b, Pythia-1.4b, and Pythia-2.8b, we set the learning rate as lr=1e-5. The weight decay is set as 1e-2. We train each model for 40 epochs. The batch size for Pythia-series model and Phi model is 16. The batch size for GPT-J-6b, Llama-2-7b, and Mistral-7b is 1. The training is based on auto-regressive loss for input sequences. For each experiment, we ran the trials five times. We report both the mean and the variance of the results to account for variability in performance.

Our experiments are conducted on A-100 machines (with memory of 80G). For four parallel GPUs, a single epoch on Phi-2-2.8b for the synthetic dataset will cost 1 hours, so totally it costs 40 hours to run on four parallel A-100 GPUs to train Phi-2-2.8b. For llama-2-7b, it costs more than 100 hours to run on four parallel GPUs to fine-tune the synthetic dataset.
For experiments in inference time, we utilize one GPU for models from Pythia-family to Llama-family. 

\subsection{Length-dependency on NTP loss}
\label{appendix:length_dependency}
\paragraph{NTP loss for conditions with varying lengths. }
Denote \( P(x_{i+1}|x_{1:i}) \) as \( P_{i+1}(x_{i+1}) \).

\[
\begin{aligned}
& \frac{\sum_{i=1}^{k+2} -\log P(y'|x_1, \ldots, x_{k+1}, x_{k+2})}{k+2} - \frac{\sum_{i=1}^{k+1} -\log P(y'|x_1, \ldots, x_{k}, x_{k+1})}{k+1} \\
= & -\frac{\log P_1(x_1)\times \dots  \times P_{k+2}(x_{k+2}) \times P_{k+3}(y') }{k+3}  + \frac{\log P_1(x_1)\times \dots  \times P_{k+1}(x_{k+1}) \times P_{k+2}(y') }{k+2} \\
= & \frac{1}{(k+3)(k+2)} \cdot \log  \frac{[ P_1(x_1)\times \dots \times P_{k+1}(x_{k+1})\times P_{k+2}(y') ]^{k+3}}{[ P_1(x_1)\times \dots \times P_{k+2}(x_{k+2})\times P_{k+3}(y') ]^{k+2}} \\
= & \frac{1}{(k+3)(k+2)} \cdot \log P_1(x_1)\times \dots \times P_{k+1}(x_{k+1}) \frac{[P_{k+2}(y')]^{k+3}}{[P_{k+2}(x_{k+2})]^{k
+2}\cdot [P_{k+3}(y')]^{k+2}}
\end{aligned}
\]

Since exploring the training dynamics of $P_i(x_i)$, $P_j(y')$ in large language models is intractable, we make a mild assumption here, at the late training stage,  $P_i(x_i)\rightarrow \hat P_i(x_i)$, $P_j(y')\rightarrow \hat P_j(y')$, in the setup with controlled variables, where samples with different lengths have same proportion of dominant conditions and suppressed conditions, then the value in log approaches $\frac{P_{k+2}(y')}{P_{k+2}(x_{k+2})}$. Since $y'$ is the false prediction made by model, whose empirical probability equals zero, so $P_{k+2}(y')$ approaches zero, then $P_{k+2}(y')< P_{k+2}(x_{k+2})$.

Given that, $\frac{P_{k+2}(y')}{P_{k+2}(x_{k+2})} < 1$

therefore,

$L_{NTP}(y'|x_{1:k+1},x_{k+2}) < L_{NTP}(y'|x_{1:k},x_{k+1})$,

we denote $L_{NTP}(y'|x_{1:k},x_{k+1})$ as $- \log \left( \frac{e^{f(\boldsymbol{x})_y}}{\sum_{y'} e^{h^{-1}(k) f(\boldsymbol{x})_{y'}} } \right)$, where $h(k)$ is positively correlated with $k$, with larger $k$ indicating larger $h(k)$.

\paragraph{Lipschitz continuity of NTP loss. }
$B_y(f)$ represents the minimal prediction on the ground truth token $y$, $i.e.$ $B_y(f):=min_{x\in S_y} f(x)_y$~\cite{wang2024unified}.

Here we prove the \textit{Lipschitz continuity}~\cite{wang2024unified} of the NTP loss, according to the definition of the NTP loss, and the above NTP loss rewriting, we have

\[
\begin{aligned}
    \mathcal{L}_\text{NTP}(f(\boldsymbol{x}), y) & = - \log \left( \frac{e^{f(\boldsymbol{x})_y}}{\sum_{y'} e^{h^{-1}(k) f(\boldsymbol{x})_{y'}} } \right) \\
    & = \log [1 + \sum_{y' \neq y} e^{h^{-1}(k) f(\boldsymbol{x})_{y'} - f(\boldsymbol{x})_y}].
\end{aligned}
\]

We denote \(\boldsymbol{s} := f(\boldsymbol{x})\), and we define

\[
\ell_y(\boldsymbol{s}) := \sum_{y' \neq y} e^{h^{-1}(k) \boldsymbol{s}_{y'}}.
\]

Therefore, we rewrite the $\mathcal{L}_\text{NTP}$ as follows:

\[
\mathcal{L}_{NTP}(f, y) = \log \left[ 1 + e^{- \boldsymbol{s}_{y}} \ell_y(\boldsymbol{s}) \right].
\]

The derivatives can be represented as follows:

\[
\begin{aligned}
    \frac{\partial \mathcal{L}_{NTP}(f, y)}{\partial \boldsymbol{s}_{y}} & = - \frac{e^{- \boldsymbol{s}_{y}} \ell_y(\boldsymbol{s})}{1 + e^{- \boldsymbol{s}_{y}} \ell_y(\boldsymbol{s})}, \\
    \frac{\partial \mathcal{L}_{NTP}(f, y)}{\partial \boldsymbol{s}_{y'}} & = h^{-1}(k) \frac{e^{- \boldsymbol{s}_{y}}}{1 + e^{- \boldsymbol{s}_{y}} \ell_y(\boldsymbol{s})} \cdot e^{h^{-1}(k) \boldsymbol{s}_{y'}}, y' \neq y. \\
\end{aligned}
\]

We can get the following inequality:

\[
\begin{aligned}
    & \Vert \nabla_{\boldsymbol{s}} \mathcal{L}_{NTP}(f, y) \Vert^2 = \left[ \ell_y(\boldsymbol{s})^2 + \sum_{y' \neq y} \left(h^{-1}(k) e^{h^{-1}(k) \boldsymbol{s}_{y'}} \right)^2 \right] \cdot \left[\frac{e^{- \boldsymbol{s}_{y}}}{1 + e^{- \boldsymbol{s}_{y}} \ell_y(\boldsymbol{s})} \right]^2 \\
    & \le \left[ \ell_y(\boldsymbol{s})^2 + \left( \sum_{y' \neq y} h^{-1}(k) \right)^2 \left( \sum_{y' \neq y} e^{h^{-1}(k) \boldsymbol{s}_{y'}} \right)^2 \right] \cdot \left[ \frac{e^{- \boldsymbol{s}_{y}}}{1 + e^{- \boldsymbol{s}_{y}} \ell_y(\boldsymbol{s})} \right]^2 \\
    & = \left[ 1 + \left( \sum_{y' \neq y} h^{-1}(k) \right)^2 \right] \cdot \left[ \frac{e^{- \boldsymbol{s}_{y}} \ell_y(\boldsymbol{s}) }{1 + e^{- \boldsymbol{s}_{y}} \ell_y(\boldsymbol{s})} \right]^2,
\end{aligned}
\]

Therefore,

\[
\begin{aligned}
    \Vert \nabla_{\boldsymbol{s}} \mathcal{L}_{NTP}(f, y) \Vert & \le \sqrt{1 + \left( \sum_{y' \neq y} h^{-1}(k) \right)^2} \frac{e^{- \boldsymbol{s}_{y}} \ell_y(\boldsymbol{s}) }{1 + e^{- \boldsymbol{s}_{y}} \ell_y(\boldsymbol{s})} \\
    & = \sqrt{1 + \left( \sum_{y' \neq y} h^{-1}(k) \right)^2} \frac{\ell_y(\boldsymbol{s})}{e^{ \boldsymbol{s}_{y}} +  \ell_y(\boldsymbol{s})} \\
    & = \sqrt{1 + \left( \sum_{y' \neq y} h^{-1}(k) \right)^2} \left[ 1 - \frac{e^{ \boldsymbol{s}_{y}}}{ \sum_{y'} e^{ h^{-1}(k) \boldsymbol{s}_{y'}} } \right] \\
    & = \sqrt{1 + \left( \sum_{y' \neq y} h^{-1}(k) \right)^2} \left[ 1 - \textit{softmax}\left( \boldsymbol{s}_{y} \right) \right].
\end{aligned}
\]

Since the score function is bounded, for any $y \in \mathcal{Y}$, there exists a constant $B_y(f)$ such that $B_y(f) = \inf_{\boldsymbol{x} \in \mathcal{S}_y} \boldsymbol{s}_y$, which completes the proof.

\subsection{More Generalized Hallucination Patterns on GPT models}

\begin{table}[htb]
\linespread{1.5}
\tabcolsep=0.036cm
\fontsize{7.8pt}{8pt}\selectfont
\caption{A wide range of hallucinations made by GPT-3.5 series of models that can be attributed to knowledge overshadowing.}
\begin{tabular}
{p{2.3cm}@{\hskip 0.2cm}p{4.0cm}@{\hskip 0.2cm}p{7cm}}
\noalign{{\color{black}\hrule height 1pt}}
Overshadowing & Input Prompt & Hallucinated Answer from LLMs  \\ \hline
\textcolor{magenta}{Not won the Nobel prize} \textbf{Overshadowed} \textcolor{cyan}{Nomination} & List some authors that were nominated but did not win the Nobel prize. & 1. Leo Tolstoy 2. James Joyce \sout{\textcolor{magenta}{3. Mark Twain}} 4. Virginia Woolf 5. Jorge Luis Borges \sout{\textcolor{magenta}{6. Emily Dickinson 7. George Orwell}} 8. Anton Chekhov 9. Jorge Amado \sout{\textcolor{magenta}{10. Thomas Hardy }}\\
\hline
\textcolor{magenta}{from Korea} \textbf{Overshadowed} \textcolor{cyan}{rock musicians} & Tell me some rock musicians from Korea. & 1. Yoon Do-hyun 2. Lee Seung-hwan 3. Shin Seung-hun 4. Seo Taiji 5. Kim Kyung-ho 6. Moon Hee-joon 7. Kang San-ae 8. Noh Soo-ram 9. Oh Seung-geun \sout{\textcolor{magenta}{10. Heo Young-saeng}} \\
\hline
\textcolor{magenta}{Description for wining the prize} \textbf{Overshadowed} \textcolor{cyan}{Negation} & Who did not receive the 1921 Nobel Prize in Physics "for his services to theoretical physics, and especially for his discovery of the law of the photoelectric effect"? & \sout{\textcolor{magenta}{German physicist Albert Einstein did not receive the 1921 Nobel}} \sout{\textcolor{magenta}{Prize in Physics for his discovery of the law of the photoelectric}} \sout{\textcolor{magenta}{effect.}} In fact, Einstein was awarded the Nobel Prize in Physics in 1921 for his explanation of the photoelectric effect, which was published in a 1905 paper titled "On a Heuristic Point of View Concerning the Production and Transformation of Light." \\
\hline
\textcolor{magenta}{AI scientist} \textbf{Overshadowed} \textcolor{cyan}{Female} & Tell me some outstanding
female AI scientist. & \sout{\textcolor{magenta}{Yoshua Bengio}}: While not female, Yoshua Bengio has significantly contributed to the advancement of AI and deep learning. As one of the "Three Musketeers of Deep Learning" alongside Geoffrey Hinton and Yann LeCun, his work has had a profound impact on the field. \\
\hline
\noalign{{\color{black}\hrule height 1pt}}
\end{tabular}
\label{tab:gpt_hallu_cases}
\end{table}


\newpage

\end{document}